\documentclass[runningheads]{templates/eccv/llncs}

\usepackage{templates/eccv/eccv}

\usepackage{templates/eccv/eccvabbrv}

\usepackage{graphicx}
\usepackage{booktabs}

\usepackage[accsupp]{axessibility}  

\usepackage{hyperref}

\usepackage{orcidlink}

\begin{document}

\title{Accelerated Spatio-Temporal Gaussian Rendering via Kinematic and Semantic Priors} 

\titlerunning{R5DGS}

\author{Denis Gridusov\inst{1}\orcidlink{0009-0001-0409-0352} \and
 Maxim Popov\inst{1}\orcidlink{0000-0003-3696-471X}\and
 Sergey Kolyubin\inst{1}\orcidlink{0000-0002-8057-1959}}

\authorrunning{D.~Gridusov et al.}

\institute{ITMO University, St. Petersburg, Russia \\ \email{	ddgridusov@itmo.ru}}
\newcommand{\name}{{R5DGS}}
\maketitle


\begin{abstract}
 Reconstructing dynamic 3D scenes from multi-view videos is a foundational task for robotics, AR/VR, and digital twins. While 3D Gaussian Splatting (3DGS) provides state-of-the art scene representation, its dynamic adaptations (4DGS) face significant challenges. Early MLP-based deformation models excel at interpolating within training frames but fail at temporal extrapolation --- important feature for forecasting in predictive systems. To address this, physically grounded approaches have been proposed to enable robust future forecasting, while, concurrently, semantic 4DGS methods have been developed for scene analysis and language-driven prompting. However, there is currently no framework that unites both physical consistency and semantic understanding. In this paper, we propose the first, to the best of our knowledge, unified 4DGS approach that is both physically grounded and semantically aware. Furthermore, to optimize computational efficiency, we introduce a novel rigid-body constraint mechanism. Instead of predicting the physical dynamics for every Gaussian particle, our model predicts the motion of a single anchor Gaussian per semantic object, updating the remaining Gaussians via relative transformations. This architectural design significantly accelerates novel view synthesis compared to an unconstrained baseline, while enabling object-accurate extrapolation and intuitive semantic interaction.
  \keywords{Semantic Scene Understanding \and Rigid-Body Dynamics \and 4D Gaussian Splatting}
\end{abstract}

\section{Introduction}
\label{sec:intro}
Reconstructing and predicting dynamic 3D scenes from multi-view video is a foundational prerequisite for a wide range of embodied applications. In robotics and autonomous navigation, accurate spatiotemporal modeling enables safe trajectory planning and proactive interaction with moving agents. In augmented or virtual reality and digital twins, it facilitates immersive scene editing and realistic predictive simulation. Beyond static view synthesis, the ability to extrapolate future states from limited observations is essential for closing the perception-action loop in unstructured environments.

Traditional 3D reconstruction has evolved from explicit discretized representations (voxels, point clouds, meshes) to implicit continuous fields such as signed distance functions~\cite{park2019deepsdf, mescheder2019occupancy} and Neural Radiance Fields (NeRF)~\cite{mildenhall2020nerf}. While NeRFs achieve high-fidelity novel view synthesis, their volumetric rendering remains computationally intensive. Recently, 3D Gaussian Splatting (3DGS)~\cite{kerbl20233d} has emerged as a dominant explicit alternative, leveraging particle-based Gaussian kernels for real-time, high-quality rendering. To capture temporal dynamics, early approaches extended NeRF by learning time-dependent deformation fields~\cite{pumarola2021d, park2021nerfies, cao2023hexplane}. These were rapidly adapted to the 3DGS paradigm, with methods like Deformable 3DGS~\cite{yang2023deformable3dgs} and 4D Gaussian Splatting~\cite{wu20244d} achieving state-of-the-art interpolation within observed video durations. However, these deformation-based approaches primarily optimize pixel-level correlations without encoding physical laws, leading to motion drift and complete failure when extrapolating beyond the training horizon.

To enable predictive modeling, recent works have integrated physical priors into 3D reconstruction. Physics-Informed Neural Networks (PINNs)~\cite{raissi2019pinns} embed partial differential equations as soft regularization terms to model phenomena like fluid dynamics~\cite{baieri2023fluid, qiu2024neusmoke} and continuum mechanics~\cite{li2023pac}. Despite theoretical guarantees, PINNs suffer from high training costs, require explicit boundary conditions or object masks, and often struggle with complex, multi-part dynamics. More recently, TRACE~\cite{li2025trace} proposes a generalized translation-rotation dynamics system per Gaussian particle, enabling label-free future extrapolation through classical mechanics and Runge-Kutta integration. Nevertheless, this per-particle formulation leads to significant computational overhead during inference and inherently lacks semantic coherence, limiting its applicability for downstream perception or control tasks.

In this work, we bridge the gap between physical dynamics, semantic controllability, and inference efficiency. We introduce \name, a unified framework that augments a physics-driven 4D Gaussian representation with instance-level semantic grouping and a rigid-body optimization scheme for extrapolation.

Our primary contributions are as follows:
\begin{itemize}
    \item A semantics-augmented physics-informed 4D Gaussian framework that enables precise Gaussian-to-object grouping via compact Identity Encodings.
    \item A centroid-driven rigid-body inference strategy that accelerates future prediction by 11 FPS while preserving trajectories plausibility, bridging the gap between high-fidelity physical extrapolation and real-time interactive deployment.
    \item An open-vocabulary querying mechanism powered by an offline CLIP-based lookup table, allowing text-driven retrieval and rendering of dynamic objects across time and viewpoints.
\end{itemize}
\section{Related Works}
\label{sec:related_works}
\subsection{Scene Representations}

The evolution of 3D scene representation has progressed from explicit geometric structures to continuous neural fields.
Early implicit approaches such as DeepSDF~\cite{park2019deepsdf} and Occupancy Networks~\cite{mescheder2019occupancy}
modeled shapes as continuous functions, enabling high-quality surface reconstruction.
Neural Radiance Fields (NeRF)~\cite{mildenhall2020nerf} further extended this paradigm by encoding volumetric radiance
as a continuous function, achieving photorealistic novel view synthesis.
However, volumetric rendering in NeRF remains computationally demanding.

To address efficiency limitations, 3D Gaussian Splatting (3DGS)~\cite{kerbl20233d}
introduced an explicit point-based representation using anisotropic Gaussians with differentiable rasterization,
achieving real-time rendering with high fidelity.
Building upon static 3D representations, dynamic extensions have emerged.
Early dynamic NeRF variants such as D-NeRF~\cite{pumarola2021d} and Nerfies~\cite{park2021nerfies}
learned time-dependent deformation fields to model motion.
Subsequent works improved efficiency using space discretization ~\cite{cao2023hexplane}.

Within the Gaussian paradigm, Deformable 3DGS~\cite{yang2023deformable3dgs}
and 4D Gaussian Splatting~\cite{wu20244d}
extended 3DGS to dynamic scenes by learning per-Gaussian deformations over time.
While these approaches achieve strong interpolation quality within observed frames,
they rely primarily on data-driven deformation models and lack explicit physical constraints,
limiting their ability to extrapolate future states.

\subsection{Semantic Gaussian Representations}
Beyond geometry and appearance, incorporating semantic understanding into 3D representations
has become increasingly important for scene interaction and editing.
For static scenes Gaussian Grouping~\cite{gaussian_grouping}
introduced learnable identity features to cluster Gaussians into object-level groups,
enabling segmentation and object-centric manipulation. Recent efforts extend the idea to dynamic settings, where object-level consistency across time is required \cite{ji2024segment}.

Concurrent work \cite{zhou2024feature} introduced foundation model distillation into 3D Gaussian Splatting by assigning a high-dimensional feature vector to each Gaussian and supervising it with pixel-aligned feature maps extracted from training images. However, since the distilled features are not inherently object-aware, decomposing the scene into meaningful instances requires post-hoc clustering in the feature space, which may lead to inconsistent object boundaries in complex scenes. Recent methods \cite{jiang2026director,li20254d} sidestep this limitation by employing multimodal object-wise prompting: a video is first converted into per-object caption features, which are then used as supervision targets. This yields representations that are naturally structured at the object level, enabling more intuitive language-guided interaction with the 3D scene, which however costs more computing resources.

However, most existing semantic Gaussian approaches focus either on static scenes
or purely deformation-based dynamics, without integrating physically grounded motion models. Consequently, they lack principled mechanisms to ensure object-level motion consistency during extrapolation.

\subsection{Physics-Consistent Dynamic Reconstruction}

Introducing physical priors into neural scene representations
has emerged as a promising direction for improving extrapolation and interpretability.

Physics-Informed Neural Networks (PINNs)~\cite{raissi2019pinns}
embed partial differential equations into the optimization objective,
and have been applied to fluid dynamics~\cite{baieri2023fluid, qiu2024neusmoke}
and continuum mechanics~\cite{li2023pac}. More recently, Physics-Informed Deformable Gaussian Splatting (PIDG)~\cite{hong2026physics}
extends this line of work to dynamic 3D Gaussian Splatting by treating each Gaussian
particle as a Lagrangian material point with time-varying constitutive parameters,
enforcing the Cauchy momentum residual as a physical constraint and supervising
particle motion via optical flow.
Although theoretically grounded, PINN-based approaches often require strong supervision, known boundary conditions, or are computationally expensive.

Alternative strategies leverage explicit simulation models. Spring-mass formulations~\cite{zhong2024reconstruction} and physics-based video generation frameworks~\cite{zhang2024physdreamer}
model specific material behaviors, while graph network simulators~\cite{sanchez2020learning}
learn particle interactions directly from data.
These approaches typically target particular object classes or material types,
limiting general applicability.

More recently, TRACE~\cite{li2025trace} proposed a Translation-Rotation Dynamics (TRD) formulation for 4D Gaussian representations. By learning per-Gaussian velocities and accelerations and integrating them via classical mechanics, TRACE enables label-free future extrapolation. However, modeling dynamics independently for each Gaussian introduces substantial computational overhead and does not provide semantic grouping.

Our work builds upon TRACE by integrating semantic-aware grouping and introducing rigid-body constraints during inference, reducing computational complexity while preserving physically consistent trajectories.



\section{Preliminary}
\paragraph{Gaussian Splatting.} 3D Gaussian Splatting~\cite{kerbl20233d} represents a static 3D scene as a set of explicit Gaussian kernels $\mathcal{G} = \{G_i\}_{i=1}^N$. Each Gaussian is parameterized by a 3D center position $\boldsymbol{\mu} \in \mathbb{R}^3$, a rotation quaternion $\mathbf{q} \in \mathbb{R}^4$, a scaling vector $\mathbf{s} \in \mathbb{R}^3$, an opacity $\alpha \in [0,1]$, and view-dependent color $\mathbf{c}$ encoded via spherical harmonics. The 3D covariance matrix is decomposed as $\mathbf{\Sigma} = \mathbf{R}\mathbf{S}\mathbf{S}^\top\mathbf{R}^\top$, where $\mathbf{R}$ and $\mathbf{S}$ are derived from $\mathbf{q}$ and $\mathbf{s}$. The 3D Gaussian kernel is evaluated as:
\begin{equation}
G(\mathbf{p}) = \exp\left(-\frac{1}{2}(\mathbf{p}-\boldsymbol{\mu})^\top \mathbf{\Sigma}^{-1} (\mathbf{p}-\boldsymbol{\mu})\right).
\end{equation}

During rendering, 3D Gaussians are projected to 2D image space. The final color of a pixel $\mathbf{p}$ is computed via $\alpha$-composition:
\begin{equation}
C(\mathbf{p}) = \sum_{i \in \mathcal{N}} T_i \alpha_i \mathbf{c}_i, \quad T_i = \prod_{j=1}^{i-1}(1-\alpha_j),
\end{equation}
where $\mathcal{N}$ is the set of Gaussians overlapping the pixel, sorted by depth.
\section{Methodology}
\label{sec:methodology}
We build upon TRACE~\cite{li2025trace}, a physics-informed 4D Gaussian framework for dynamic scene extrapolation. Given multi-view RGB videos, TRACE represents the scene at a canonical timestamp $t=0$ as a set of 3D Gaussians $\mathcal{G}^0 = \{(\mathbf{x}_i, \mathbf{r}_i, \mathbf{s}_i, \alpha_i, \mathbf{c}_i)\}_{i=1}^N$.

Temporal evolution is modeled via two parallel modules: (1) an auxiliary deformation field $f_{\text{def}}$ that predicts per-Gaussian displacements $(\Delta\mathbf{x}, \Delta\mathbf{r}, \Delta\mathbf{s})$ for interpolation within observed time, and (2) a Translation-Rotation Dynamics (TRD) module that learns physical parameters: equivalent center velocity $\bar{\mathbf{v}}_c$, acceleration $\bar{\mathbf{a}}_c$, angular velocity $\boldsymbol{\omega}_p$, and angular acceleration $\boldsymbol{\varepsilon}_p$ to enable future frame extrapolation via 2nd-order Runge-Kutta integration. While TRACE achieves strong extrapolation quality, it treats each Gaussian independently, resulting in high inference cost and lacking semantic object awareness.
\subsection{Identity-Augmented Representation}

To enable object-level reasoning, we augment each 3D Gaussian with a compact, learnable 16-dimensional identity vector $\mathbf{e}_i \in \mathbb{R}^{16}$, inspired by Gaussian Grouping~\cite{gaussian_grouping}. During differentiable rendering, these features are $\alpha$-blended to the image plane:
\begin{equation}
\mathbf{E}_{\text{id}} = \sum_{i \in \mathcal{N}} \mathbf{e}_i \alpha'_i \prod_{j=1}^{i-1} (1-\alpha'_j),
\end{equation}
where $\alpha'_i$ denotes the projected opacity. The rendered identity map is supervised by:
\begin{itemize}
    \item {2D Identity Loss.} Classifier $f$ maps $\mathbf{E}_{\text{id}}$ to $\mathbf{K}$ object classes, optimized via cross-entropy against coherent multi-view masks.
    \item {3D Spatial Regularization.} A KL-divergence penalty enforces feature consistency among $k$-nearest spatial neighbors in canonical space, mitigating supervision gaps from occlusions.
\end{itemize}

These semantic priors are jointly optimized with standard photometric reconstruction, yielding the total objective:
\begin{equation}
\label{eq:5dgs}
\mathcal{L}_\text{5DGS} = \mathcal{L}_{\text{render}} + \lambda_{\text{obj}}\mathcal{L}_{\text{obj}} + \lambda_{\text{3d}}\mathcal{L}_{\text{3d}},
\end{equation}
where $\mathcal{L}_{\text{render}} = (1-\lambda_{\text{dssim}})\mathcal{L}_1 + \lambda_{\text{dssim}}(1-\text{SSIM})$ following the 3DGS formulation~\cite{kerbl20233d}, $\mathcal{L}_{\text{obj}}$ is the 2D Identity Loss, $\mathcal{L}_{\text{3d}}$ is the 3D Spatial Regularization and $\lambda_{\text{obj}}, \lambda_{\text{3d}}$ are scalars.

This gives a discrete, by-design grouping of Gaussians into object instances while preserving reconstruction quality.

\subsection{Rigid-Body Constrained Extrapolation}
\label{sec:rigid_extrap}
The TRD module in TRACE predicts dynamics and integrates position for all scene Gaussians, which is computationally expensive during extrapolation ($t > t_{\max}$). From classical mechanics, we know that the motion of particles within a rigid body is strictly governed by constraints that maintain constant relative distances. Guided by this principle, we introduce a group-level rigid-body constraint during inference, substituting per-Gaussian dynamics with physically consistent object-level propagation.

Let $\mathcal{G}_k$ denote Gaussians assigned to object $k$. We select a representative Gaussian $r_k \in \mathcal{G}_k$ as the one closest to the geometric centroid and precompute canonical offsets:
\begin{equation}
\mathbf{o}_i = \mathbf{x}_i - \mathbf{x}_{r_k}, \quad \forall i \in \mathcal{G}_k.
\end{equation}
During extrapolation, we first propagate all Gaussians to $t_{\max}$ using $f_{\text{def}}$. We then integrate TRD dynamics only for the $K$ representatives, obtaining future position $\mathbf{x}_{r_k}^{\text{vel}}$ and orientation quaternion $\mathbf{q}_{r_k}^{\text{vel}}$. Motion is rigidly propagated to remaining Gaussians via:
\begin{align}
\Delta\mathbf{x}_i &= \bigl(\mathbf{x}_{r_k}^{\text{vel}} - \mathbf{x}_{r_k}^{\text{def}}\bigr) + \bigl(\mathbf{R}(\mathbf{q}_{r_k}^{\text{vel}})\mathbf{o}_i - \mathbf{o}_i\bigr), \label{eq:rigid_prop} \\
\mathbf{q}_i^{\text{out}} &= \mathbf{q}_{r_k}^{\text{vel}} \otimes \Delta\mathbf{q}_i^{\text{def}},
\end{align}
where $\mathbf{R}(\cdot)$ converts quaternions to rotation matrices, $\otimes$ denotes Hamilton product, and superscript $def$ indicates deformation network outputs. This formulation preserves inter-point distances and relative orientations by construction while reducing MLP and integrator queries from $\mathcal{O}(N)$ to $\mathcal{O}(K)$, where $K \ll N$.

\subsection{Additional Loss Components}
To increase reconstruction quality we introduce several loss components.
\paragraph{Rigid Distance Preservation.} To encourage rigid motion during training, we introduce:
\begin{equation}
\label{eq:l_rigid}
\mathcal{L}_{\text{rigid}} = \frac{1}{N}\sum_{i=1}^{N} \Bigl( \bigl\|\mathbf{x}_i^{\text{after}} - \mathbf{x}_{r_{g(i)}}^{\text{after}}\bigr\| - \bigl\|\mathbf{x}_i^{\text{before}} - \mathbf{x}_{r_{g(i)}}^{\text{before}}\bigr\| \Bigr)^2,
\end{equation}
where $\mathbf{x}_i^{\text{before}}$ denotes the Gaussians center position before applying the predicted by $f_\text{def}$ and TRD displacement  and $\mathbf{x}_i^{\text{after}}$ are those  positions after deformation. The loss penalizes changes in distance from each Gaussian to its object representative. This soft constraint complements the hard rigid propagation at inference.

\paragraph{Semantic Majority Consistency.} For maintaining semantic coherence, we found that soft KL-divergence regularization is not enough, so add extra penalty to minimize MSE between a query Gaussian's predicted class distribution ($p_q$) and the mean of its $k$ spatial neighbors ($p_n$):
\begin{equation}
\label{eq:l_major}
\mathcal{L}_{\text{major}} = \frac{1}{M}\sum_{q=1}^{M} \Bigl\| \mathbf{p}_q - \frac{1}{k}\sum_{n \in \mathcal{N}_k(q)} \mathbf{p}_n \Bigr\|^2.
\end{equation}

Losses $\mathcal{L}_{\text{3d}}$ and $\mathcal{L}_{\text{major}}$ are computed every $\tau_{\text{reg}}$ iterations, $\mathcal{L}_{\text{rigid}}$ activates after iteration $t_{\text{rigid}}$ once the deformation field stabilizes.

\subsection{CLIP-like Open-Vocabulary Querying}
To enable text-driven scene interaction, we construct an offline lookup table. For each object group $g \in \{1,\ldots,K\}$, we extract a representative masked view, encode it with CLIP-like~\cite{radford2021learning} model, and store the text-aligned embedding $\mathbf{t}_g$. At inference, a natural language prompt $\mathbf{q}_{\text{text}}$ is encoded via the same text encoder, and object groups are retrieved by maximizing cosine similarity:
\begin{equation}
g^* = \arg\max_{g} \cos\bigl(\text{CLIP}_{\text{text}}(\mathbf{q}_{\text{text}}), \mathbf{t}_g\bigr).
\end{equation}
The retrieved Gaussian subset $\mathcal{G}_{g^*}$ can be independently rendered at arbitrary timestamps $t$ and camera poses $\mathbf{C}$, supporting object isolation, editing, and selective visualization without retraining.
\section{Experiments}
\subsection{Implementation Details}
For data preparation, we employ SAM2 \cite{ravi2024sam2} combined with DEVA \cite{cheng2023tracking} tracking to generate consistent object masks across all views and timestamps. This ensures temporal and multi-view consistency required for training identity encoding vectors. We benchmark our method on the Dynamic Indoor Scene dataset~\cite{li2023nvfi}, which contains four scenes with multiple objects undergoing independent rigid body motions.

Our training configuration includes the following hyperparameters: for 3D semantic regularization, we set $k=5$ nearest neighbors with $\lambda_{\text{3d}}=2.0$, computed every 2 iterations on samples of 1000 points (from maximum 300,000). For majority consistency loss, we use $k=5$ neighbors with $\lambda_{\text{maj}}=0.5$, computed on 1000 sampled points. The rigid distance loss is activated after iteration 10,000 with $\lambda_{\text{rigid}}=0.5$.
\subsection{Notation and Method Variants.}
To systematically evaluate the contribution of each component, we introduce the following naming convention. Each variant is defined by two independent choices: (1) the set of loss functions used during {training}, and (2) the extrapolation strategy applied during {inference}.

\begin{table}[h]
    \centering
    \small
    \renewcommand{\arraystretch}{1.2}
    \caption{Summary of method variants. All variants share the same Identity Encoding architecture and CLIP-based lookup table.}
    \begin{tabular}{l|c|c}
    \toprule
    \textbf{Variant} & \textbf{Training losses} & \textbf{Inference}  \\
    \midrule
    {5DGS} & $\mathcal{L}_{\text{5DGS}}$ & Standard  \\
    {R5DGS} & $\mathcal{L}_{\text{5DGS}}$ & Rigid-body \\
    {R5DGS w/ extra loss} & $\mathcal{L}_{\text{5DGS}} + \mathcal{L}_{\text{rigid}} + \mathcal{L}_{\text{major}}$ & Rigid-body \\
    \bottomrule
    \end{tabular}
    \label{tab:variants}
\end{table}
Based on the loss function we use, we define three method variants evaluated in our experiments, as shown in Table \ref{tab:variants}.

Inference strategies:
\begin{itemize}
    \item Standard: per-Gaussian dynamics prediction via the full TRD module (as in TRACE~\cite{li2025trace}).
    \item Rigid-body: group-level dynamics prediction for representative Gaussians only, with rigid propagation to the rest of the object (Sec.~\ref{sec:rigid_extrap}).
\end{itemize}

\subsection{Benchmark Results}
We evaluate our method across theree key dimensions: rendering quality (PSNR, SSIM, LPIPS), segmentation accuracy (mIoU) and novel view synthesis speed (FPS) on unseen timestamps (extrapolation).
\begin{table*}[h]
    \centering
    \small
    \setlength{\tabcolsep}{3pt}
    \renewcommand{\arraystretch}{1.3}
    \caption{mIoU and FPS results for different model variants to validate rigid constraints effect. The best results are denoted with \textbf{bold}. FPS is calculated on NVIDIA RTX 4090}
    \begin{tabular}{l|cc|cc|cc|cc|cc}
        \toprule
         & \multicolumn{2}{c|}{Dining Table} 
         & \multicolumn{2}{c|}{Chessboard} 
         & \multicolumn{2}{c|}{Darkroom} 
         & \multicolumn{2}{c|}{Factory}
         & \multicolumn{2}{c}{Overall}\\ \cline{2-11}
        Methods 
        & FPS$\uparrow$ & mIoU$\uparrow$
        & FPS$\uparrow$ & mIoU$\uparrow$
        & FPS$\uparrow$ & mIoU$\uparrow$
        & FPS$\uparrow$ & mIoU$\uparrow$ 
        & FPS$\uparrow$ & mIoU$\uparrow$ \\
        
        \midrule
        5DGS
        & 66.9 & \textbf{0.78} 
        & 67.3 & \textbf{0.75}
        & 49.4 & 0.37 
        & 64.9 & \textbf{0.47} 
        & 62.1 & \textbf{0.59}\\
        
        R5DGS
        & \textbf{76.3} & 0.77 
        & \textbf{76.9} & 0.73 
        & \textbf{66.2} & \textbf{0.38}
        & \textbf{75.0} & 0.46
        & \textbf{73.6} & \textbf{0.59}\\
        \bottomrule
    \end{tabular}
    \label{tab:results_fps}
\end{table*}
\begin{table*}[h]
    \centering
    \small
    \setlength{\tabcolsep}{3pt}
    \renewcommand{\arraystretch}{1.3}
    \caption{Reconstruction metric results for different model variants compared with TRACE \cite{li2025trace}.}
    \resizebox{\linewidth}{!}{%
    \begin{tabular}{l|ccc|ccc|ccc|ccc}
        \toprule
         & \multicolumn{3}{c|}{Dining Table} 
         & \multicolumn{3}{c|}{Chessboard} 
         & \multicolumn{3}{c|}{Darkroom} 
         & \multicolumn{3}{c}{Factory} \\ \cline{2-13}
        Methods 
        & PSNR$\uparrow$ & SSIM$\uparrow$ & LPIPS$\downarrow$
        & PSNR$\uparrow$ & SSIM$\uparrow$ & LPIPS$\downarrow$
        & PSNR$\uparrow$ & SSIM$\uparrow$ & LPIPS$\downarrow$
        & PSNR$\uparrow$ & SSIM$\uparrow$ & LPIPS$\downarrow$ \\
        
        \midrule
        \midrule


        TRACE 
        & 35.580 & 0.962 & 0.0497 
        & 34.630 & 0.963 & 0.055 
        & 37.774 & 0.961 & 0.067 
        & 36.488 & 0.965 & 0.049 \\

        5DGS (Ours)
        & 35.428 & 0.956 & 0.055 
        & 33.991 & 0.956 & 0.063 
        & 36.600 & 0.955 & 0.074 
        & 35.926 & 0.958 & 0.055 \\

        \midrule

        R5DGS (Ours) 
        & \textbf{28.844} & \textbf{0.942} & \textbf{0.066} 
        & 28.805 & \textbf{0.932} & \textbf{0.086} 
        & 31.181 & 0.939 & 0.091 
        & 29.798 & 0.924 & 0.075 \\

        R5DGS w/ extra loss (Ours) 
        & 28.688 & 0.939 & 0.067 
        & \textbf{29.153} & 0.929 & 0.089 
        & \textbf{31.537} & \textbf{0.943} & \textbf{0.087} 
        & \textbf{30.749} & \textbf{0.928} & \textbf{0.073} \\

        \bottomrule
    \end{tabular}
    }
    \label{table:results_render_quality}
\end{table*}
\begin{figure*}[!htb]
    \centering
    \includegraphics[width=1\textwidth]{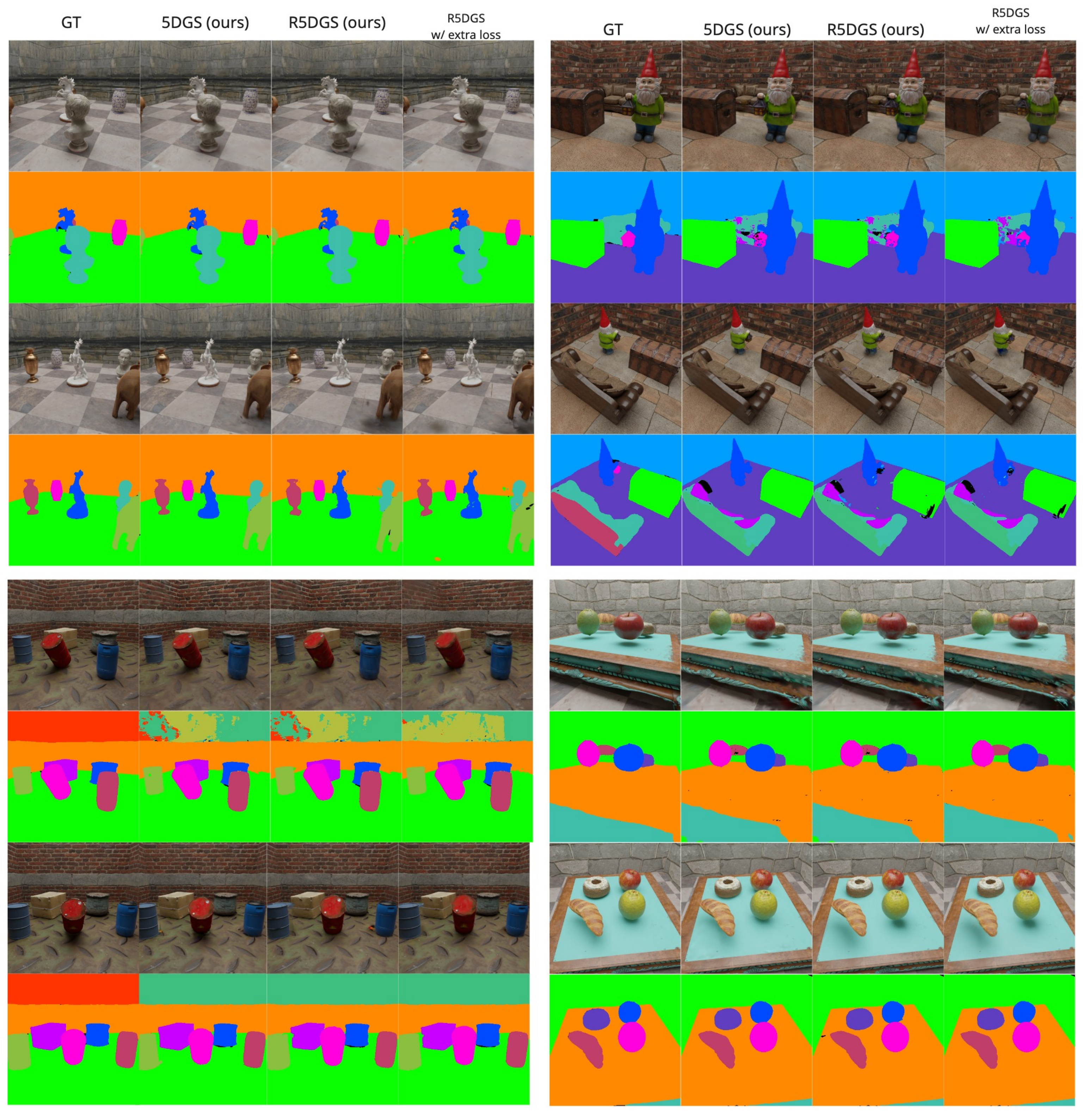}
    \caption{Visual comparison of different model variants with ground truth rgb and semantics.}
    \label{fig:all_viz}
\end{figure*}

\paragraph{Results Analysis.}

Table~\ref{tab:results_fps} reports the runtime and segmentation
accuracy of each variant, while Table~\ref{table:results_render_quality}
evaluates the photometric rendering quality.By restricting physics prediction to a small set of representative Gaussians per object (typically 9-12 for the tested dataset) rather than the full set ($\sim$40,000), our approach achieves a consistent 11 FPS speedup during extrapolation novel view synthesis. While this group-level constraint introduces a reduction in photometric fidelity, mainly stemming from a small number of misclassified Gaussians near object boundaries and the exclusion of shadows from trajectory propagation, as they are not classified as part of the object, it preserves physically plausible motion trajectories and maintains structural coherence across future frames, as visualized in Figure~\ref{fig:all_viz}. Furthermore, segmentation accuracy remains nearly unaffected, confirming that the rigid propagation effectively preserves object-level structure. 

\paragraph{Segmentation Failures Analysis.}
We observe notably lower mIoU on Darkroom and Factory scenes. This is primarily attributed to tracking errors during the offline mask preparation stage with SAM2+DEVA, where occlusions and rapid motions caused identity switches. Figure~\ref{fig:all_viz} visualizes representative failure cases where the tracker loses object consistency across frames.
\subsection{Open-Vocabulary Grounding}
\begin{figure}[htbp]
    \centering
    \includegraphics[width=1\linewidth]{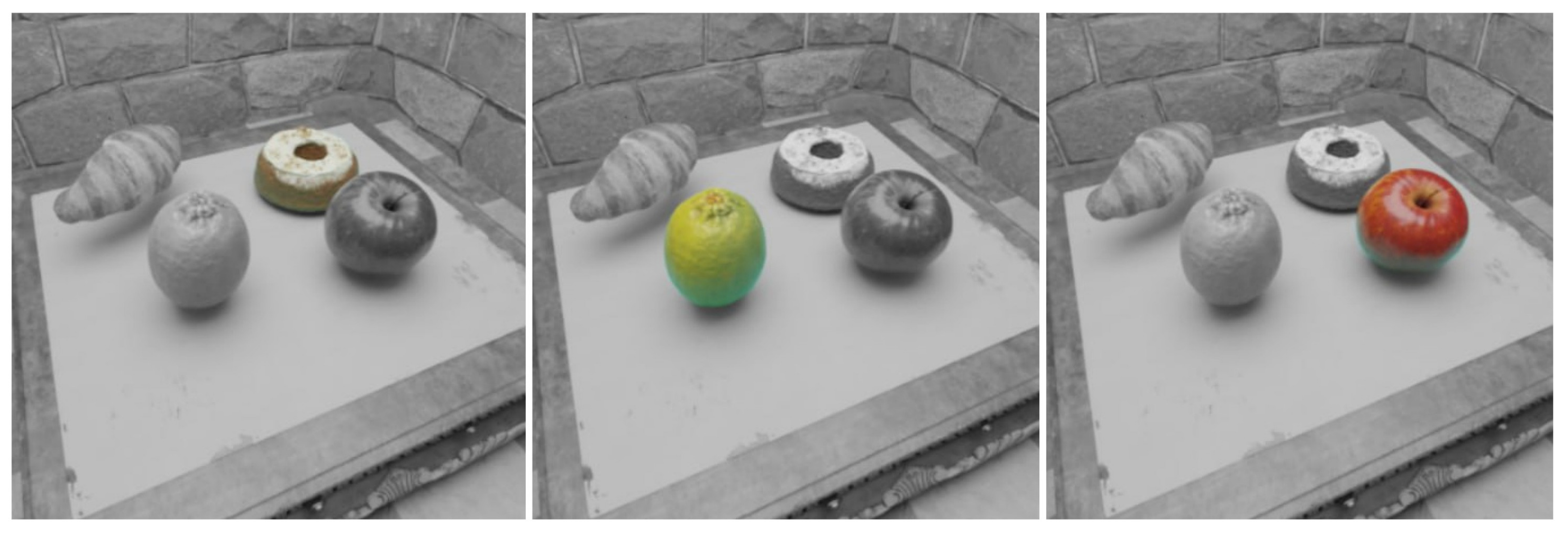}
    \caption{Grounding result visualization with prompts ``donut'', ``lemon'' and ``apple''.}
    \label{fig:grounding_demo}
\end{figure}
Beyond reconstruction and segmentation, our framework enables semantic interaction with the 4D scene. By constructing an offline CLIP-based lookup table that maps object Gaussian groups to text embeddings, we support natural language queries for object retrieval and selective rendering. Specifically, we used Perception Encoder \cite{bolya2026perception} for extracting image and text embeddings. Figure~\ref{fig:grounding_demo} demonstrates successful grounding of queries like ``donut'' and ``apple'', where the system retrieves the corresponding Gaussian subset and renders it at arbitrary timestamps and viewpoints. This capability opens new possibilities for robot instruction following and interactive scene editing without retraining.
\section{Conclusion}
In this work, we introduced a semantics-augmented, physics-informed 4D Gaussian framework that bridges the gap between high-fidelity dynamic scene extrapolation and real-time interactive deployment. By integrating compact Identity Encodings with a Translation-Rotation Dynamics system, we enable precise Gaussian-to-object grouping and open-vocabulary text-driven querying via an offline CLIP lookup table. Crucially, our rigid-body inference constraint replaces per-Gaussian physics prediction with centroid-driven integration and rigid propagation, yielding an 11 FPS speedup during extrapolation while preserving physically plausible motion trajectories and structural coherence. 

Despite these advances, the rigid-body assumption introduces a moderate trade-off in photometric fidelity, primarily affecting highly deformable regions or scenes with imperfect mask tracking. The current pipeline also relies on offline SAM2+DEVA mask association, which can suffer from identity switches under severe occlusions or rapid motion. Future work will focus on enhancing the semantic supervision and extending our rigid-body formulation to articulated and composite objects. 

Finally, our approach demonstrates that semantic-aware, physics-constrained Gaussian representations offer a practical path toward efficient, controllable 4D scene understanding for embodied applications.



%
%
\bibliographystyle{splncs04}
\bibliography{main}
\end{document}